# Multimodal registration of FISH and nanoSIMS images using convolutional neural network models

Xiaojia He[1,†] (xiaojia.he@emory.edu), Christof Meile[1] (cmeile@uga.edu), Suchendra M. Bhandarkar[2] (suchi@uga.edu)

[1]Department of Marine Sciences, University of Georgia, Athens, GA, USA

[2]Department of Computer Science, University of Georgia, Athens, GA, USA

## Abstract

Nanoscale secondary ion mass spectrometry (nanoSIMS) and fluorescence in situ hybridization (FISH) microscopy provide high-resolution, multimodal image representations of the identity and cell activity respectively of targeted microbial communities in microbiological research. Despite its importance to microbiologists, multimodal registration of FISH and nanoSIMS images is challenging given the morphological distortion and background noise in both images. In this study, we use *convolutional neural networks* (CNNs) for multiscale feature extraction, *shape context* for computation of the minimum transformation cost feature matching and the *thin-plate spline* (TPS) model for multimodal registration of the FISH and nanoSIMS images. Registration accuracy was quantitatively assessed against manually registered images, at both, the pixel and structural levels using standard metrics. Although all six tested CNN models performed well, *ResNet18* was observed to outperform *VGG16*, *VGG19*, *GoogLeNet* and *ShuffleNet* and *ResNet101* based on most metrics. This study demonstrates the utility of CNNs in the registration of multimodal images with significant background noise and morphology distortion. We also show aggregate shape, preserved by binarization, to be a robust feature for registering multimodal microbiology-related images.

[†] Present affiliation: Division of Pulmonary, Allergy and Critical Care Medicine, Emory University School of Medicine, Atlanta, GA, USA

## Introduction

Nanoscale secondary ion mass spectrometry (nanoSIMS) is a powerful tool to quantify elemental distribution at nanometer-scale resolution (1). Combining nanoSIMS imaging with fluorescence in situ hybridization (FISH) microscopy allows one to study microbial activity and correlate it with the identity of cells (2). However, he nanoSIMS and FISH images display unequal magnification and distortion. Several image registration algorithms exploit geometrical information to align the input images (3). Notably, feature-based registration methods rely on point- or shape-based correspondences between two images where the features, such as corners or contours of structures, are either derived automatically from the underlying image or from markers with known positions. Once the corresponding points are selected, their locations in the two images are used to reconstruct a spatial transformation (4, 5). In contrast, in intensity-based methods, only pixel intensity values, instead of specific features, are considered to determine the spatial transformation.

Deep learning has been increasingly recognized as a powerful toolbox for multimodal image registration, especially in medical imaging (6, 7) and remote sensing (8, 9). The *convolutional neural network* (CNN) is a widely used deep neural network (DNN) architecture comprising of convolutional layers, max-pooling layers and a softmax layer, in addition to problem-specific layers. The CNN has been used extensively for feature extraction in image classification (10, 11), image segmentation (12, 13) and image registration (14, 15) tasks, and several variants of the CNN architecture have been proposed for multimodal image registration (6, 16, 17). In the past decade, CNNs have been shown to be very successful in solving biomedical multimodal image registration problems (18-21). Notably, several CNN architectures have won awards in international competitions such as the ImageNet Large Scale Visual Recognition Challenge (ILSVRC), for example GoogLeNet (ILSVRC 2014 winner) (22) and VGG16 (ILSVRC 2014 runner up) (23). Additionally, one can easily leverage existing, CNN architectures that are well recognized for their high accuracy and speed in image classification and feature extraction, with pretrained weights derived from the several million training images in the ImageNet database (http://www.image-net.org). Although images of neither microorganisms nor microbial aggregates are available in the ImageNet database, deep CNN architectures that are pre-trained on ImageNet are excellent at general image feature extraction. To the best of our knowledge, this is the first documented application of deep CNN models to extract features from and subsequently register multimodal microbial images. Here, we present an automated scheme to register FISH and nanoSIMS images using multiple CNN models. The convolutional feature map is extracted at multiple image resolutions and used for feature point selection. *Shape context* is used to identify matched features and the *thin-plate spline* (TPS) model is employed to register the FISH and nanoSIMS images by computing a transformation matrix. The results using the different CNNs, feature matching approaches and transformation computation and registration methods are compared and discussed.

## Methods

FISH and nanoSIMS images were acquired by McGlynn et al. (24). Image analysis and registration accuracy measurement methods with manual image alignment are described in (24). In this work, both raw RGB images and their binarized versions were used. The computational workflow for the multimodal registration scheme is depicted in Fig. 1. Briefly, the input images were preprocessed to remove background noise and then fed to selected CNN models with pretrained weights. Features were then extracted at desired predetermined layer depths (scales) using the selected CNN models. A subset of the extracted features was selected and further constrained to generate a 2-D array of matched feature points using *shape context* and *bipartite graph matching* algorithms. Finally, the matched feature points were used for image transformation computation and image registration using the TPS model. Quantitative registration accuracy metrics such as the *root mean squared error* (RMSE), *structural similarity index* (SSIM), and *average absolute intensity difference* (AAID) were computed at both, the pixel and structural levels. The details of each workflow stage are explained in the following subsections.

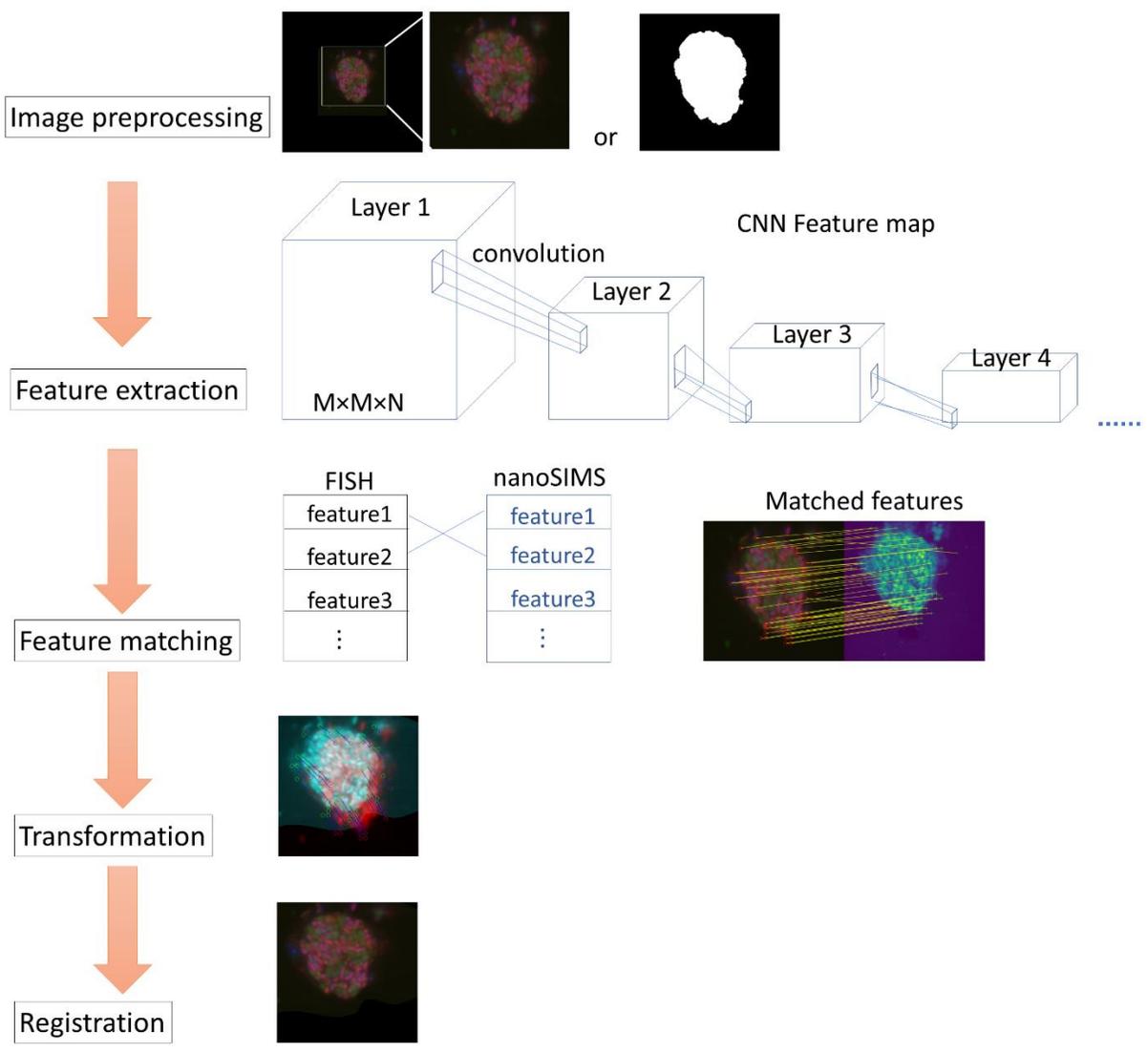

**Fig. 1.** Workflow for multimodal registration of FISH and nanoSIMS images. Use grayscale image in print.

**Non-rigid registration using a CNN**

CNN architectures *ShuffleNet* (25), *GoogLeNet* (22), *ResNet-18* and *ResNet-101* (26), and *VGG16* and *VGG19* (23) were used in this study, with pretrained weights derived from the several million training images in the ImageNet database (http://www.image-net.org). Input images to the CNN were either raw RGB or preprocessed binary FISH and nanoSIMS images (see supplement A1). All input images were rescaled to a size of 224×224 pixels and fed through the convolutional layers. Feature points were extracted from the FISH and nanoSIMS images at multiple CNN layers and used for multimodal registration via TPS interpolation.

*Feature points extraction and matching*

Features were extracted from the final layer of each individual module in the CNN architecture starting with a layer size of 28×28, and proceeding to 14×14 and 7×7, as indicated in Figs. S2 – S7. The selection of convolutional layers was heuristic and aimed to include both high- and low-level features. The feature map obtained from each layer was normalized by applying the transformation $z = (x - \mu)/\sigma$, where the feature points $x$ in each feature map are assumed to be normally distributed with mean $\mu$ and standard deviation $\sigma$. Next, we generated the feature distance map by computing the symmetric matrix of pairwise feature distance values. We concatenated the feature distance maps from each layer to yield a single feature distance map for each FISH and nanoSIMS image pair. We then processed the concatenated feature distance map by selecting the smallest value from each row and using the match threshold to select the top 20% matched points.

*Shape context*

After selecting the preliminary matching features, we used *shape context* to determine the matching that minimizes a transformation cost function that quantifies shape similarity based on the neighborhood structure of a feature point on a shape contour. The shape context at feature point $p_i$ is defined as a histogram $h_i$ of the relative coordinates $q$ of the remaining $n-1$ feature points (27):

$$h_i(k) = \#\{q \neq p_i : |(q - p_i)| \in bin(k)\} \qquad \text{Eq. 1}$$

where the bins uniformly divide the log-polar space. To obtain a shape context descriptor, we computed the distance values between points in the matched feature map and normalized them by the mean. Next, we computed the shape context descriptor by directly counting the points within each radial and angular region as previously assigned.

*Bipartite graph matching*

Consider minimizing the total cost of matching given by

$$H(\pi) = \sum_i C(p_i, q_{\pi(i)}) \qquad \text{Eq. 2}$$

where $\pi$ denotes a permutation, and $C$ is the cost function defined as $C_{i,j} = 1/2 \sum_{k=1}^{K} \frac{(h_i(k) - h_j(k))^2}{h_i(k) + h_j(k)}$, where $h_i$ and $h_j$ are the obtained shape context descriptors (normalized $k$-bin histograms) for the matched feature points $p_i$ and $q_j$ on the FISH and nanoSIMS images, respectively. The resulting weighted bipartite graph matching problem based on $H(\pi)$ was solved using the efficient Jonker-Volgenant algorithm (28). Finally, we computed the Euclidian distance between each matched feature pair and only retained the matches that fall between the 25% and 75% quantile as inliers. The values of the matching threshold were chosen based on trial and error.

*Transformation and Registration*

Given a finite set of point correspondences between two shapes, image transformation and registration function $T: \mathbb{R}^2 \to \mathbb{R}^2$ can be realized using the TPS model (29) which performs non-rigid registration or alignment of deformed images. The underlying transformation was modeled as a radial-basis function where the foreground pixels of the moving image deform under the influence of the control points $p_i$, $i = 1, \ldots n$.

*Similarity registration*

Similarity registration was used as a comparison to our proposed non-rigid, TPS-based registration scheme. Using the features extracted from CNN models, similarity registration allows for alignment of images via a combination of translation, rotation, and scaling parameters (30, 31).

*Comparison to a state-of-art registration method and non-CNN feature extraction methods*

A state-of-art registration scheme, termed as the CoMIR (Contrastive Multimodal Image Representations) method proposed by Pielawski et al., 2021 (40) was implemented in this study for comparison, following the directions from GitHub repository (https://github.com/MIDA-group/CoMIR). The CoMIR approach showed outstanding performance in the registration of remote sensing images and biomedical images, outperforming several other state-of-the-art image registration methods. To further evaluate the performance of our CNN-based methods, we also implemented and evaluated a variety of non-CNN-based feature extraction methods such as SURF (Speeded Up Robust Features, Bay et al. 2008(35)), KAZE (Alcantarilla et al. 2012(36)), BRISK (Binary Robust Invariant Scalable Keypoints, Leutenegger et al. 2011(37)), Harris (Harris et al. 1988(38)), and FAST (Features from Accelerated Segment Test, Rosten et al. 2005(39)) features (Fig. S15 and Fig. S16). SURF is a similarity-invariant, fast and robust algorithm for local feature extraction (35). KAZE is a scale- and rotation-invariant, fast multiscale feature detection and description approach for nonlinear scale spaces (36). BRISK is a scale- and rotation-invariant, fast feature point extraction algorithm (37). In contrast, Harris (38) and FAST (39) are corner detection algorithms commonly used to extract corner-like features in an image.

*Quantitative image registration assessment*

Automated registration was compared to manually registered images (the ground truth). Three different error metrics were employed to assess registration accuracy at the pixel and structural levels: *root mean squared error* (RMSE), *structural similarity index* (SSIM), and *average absolute intensity*

*difference* (AAID). RMSE quantifies the difference between registered images $(\hat{y}, y)$ by computing the square root of the mean square error of pixel values over RGB channels between two images (38):

$$RMSE = \sqrt{\frac{\sum_{i=1}^{N}(\hat{y}_i - y_i)^2}{N}} \qquad \text{Eq. 3}$$

SSIM measures the perceived similarity in structural information between two images and entails computing a weighted combination of the *luminance index l*, the *contrast index c* and the *structural index s* (39):

$$SSIM = [l(\hat{y}, y)]^\alpha [c(\hat{y}, y)]^\beta [s(\hat{y}, y)]^\gamma \qquad \text{Eq. 4}$$

with $l(\hat{y}, y) = \frac{2\mu_{\hat{y}}\mu_y + c_1}{\mu_{\hat{y}}^2 + \mu_y^2 + c_1}$, $c(\hat{y}, y) = \frac{2\sigma_{\hat{y}}\sigma_y + c_2}{\sigma_{\hat{y}}^2 + \sigma_y^2 + c_2}$, and $s(\hat{y}, y) = \frac{\sigma_{\hat{y}y} + c_3}{\sigma_{\hat{y}}\sigma_y + c_3}$, where $\mu_{\hat{y}}$, $\mu_y$ are the local means; $\sigma_{\hat{y}}$, $\sigma_{\hat{y}}$ the standard deviations; and $\sigma_{\hat{y}y}$ the cross-covariance for images $\hat{y}$ and $y$, respectively. The weights $\alpha$, $\beta$ and $\gamma$ were set to 1.

The AAID metric is based on the absolute intensity difference between two images (32):

$$AAID = \frac{1}{MNQ} \sum_{i=1}^{M} \sum_{j=1}^{N} \sum_{k=1}^{Q} |\hat{y}_{i,j,k} - y_{i,j,k}| \qquad \text{Eq. 5}$$

where $M$, $N$, and $Q$ represent the dimension of images.

Smaller RMSE and AAID values represent a better registration result, whereas SSIM is larger for better aligned images.

## Results and Discussion

Visual (Figs. 2 and 3) and quantitative (Table 1) comparison of our results with manually registered images shows good agreement, signifying the advantages of the automated registration. Image preprocessing with binary thresholding (see Supplemental A1) significantly improved the accuracy of image registration compared to RGB images (left panels in Figs. 2 and 3). Binarization yielded substantially better results than when analyzing the raw RGB images (Table 1), i.e., the pixel differences in RMSE and AAID were very small (TPS registration, Table 1), whereas the SSIM index was > 0.8 for a significantly deformed FISH image and > 0.78 for a deformed FISH image with multiple components (TPS registration). This shows that the aggregate shape, which is deformed but consistent between image modalities, is critical. The addition of intra-aggregate features, which may differ between the FISH and nanoSIMS images, results in deterioration of the registration results (Figs. 4 and 5). It is also noted that residual component(s) in the FISH and nanoSIMS images outside the region of interest (ROI) did not match well even after exhaustive trial and error iterations (see binary images in Figs. 2 and 3).

However, mismatches between small components due to the binarization preprocessing did not impact registration of our microbial aggregate images. Although a dominant object exists in the image examined (Fig. 5), there is no need to first remove other smaller components in the two images before proceeding to align them. This is largely due to the observation that features extracted using the CNNs were mostly found to be from the dominant object in the image (Fig. 5).

Our results show that TPS-based registration outperforms registration based on similarity metrics (Figs. 2 and 3). With radial basis functions, TPS-based registration is capable of locally transforming and warping the target FISH image onto the nanoSIMS image. In contrast, similarity-based registration entails only global linear transformations, i.e., rotation, scaling, and translation (33), leading to significant disparity in registration results between TPS-based and similarity-based registration (Fig. 2). While computationally more intensive, TPS-based registration introduces smooth, elastic deformations, producing a reasonably well-aligned image even for a significantly deformed FISH image (Fig. 2). This finding is consistent with the reported high accuracy and robustness of TPS in data interpolation and image registration (34).

Although there are no images of microorganisms or microbial aggregates in the ImageNet database, our results indicate that all six CNN models yield high registration accuracy at both, the pixel and structural levels (Figs. 2 and 3, Table 1). The analysis of a significantly deformed image (Fig. 2), and a deformed image with multiple components (Fig. 3) revealed that, in general, features identified by *ResNet* and *ShuffleNet* were more complex than their *VGG* and *GoogLeNet* counterparts (Figs. S8 and S10), potentially contributing to slightly better registration results for a significantly deformed FISH image (Fig. 2 and Table 1). Moreover, CNN models also performed well in more complicated scenarios comprising of a deformed FISH image with multiple components (Fig. 3).

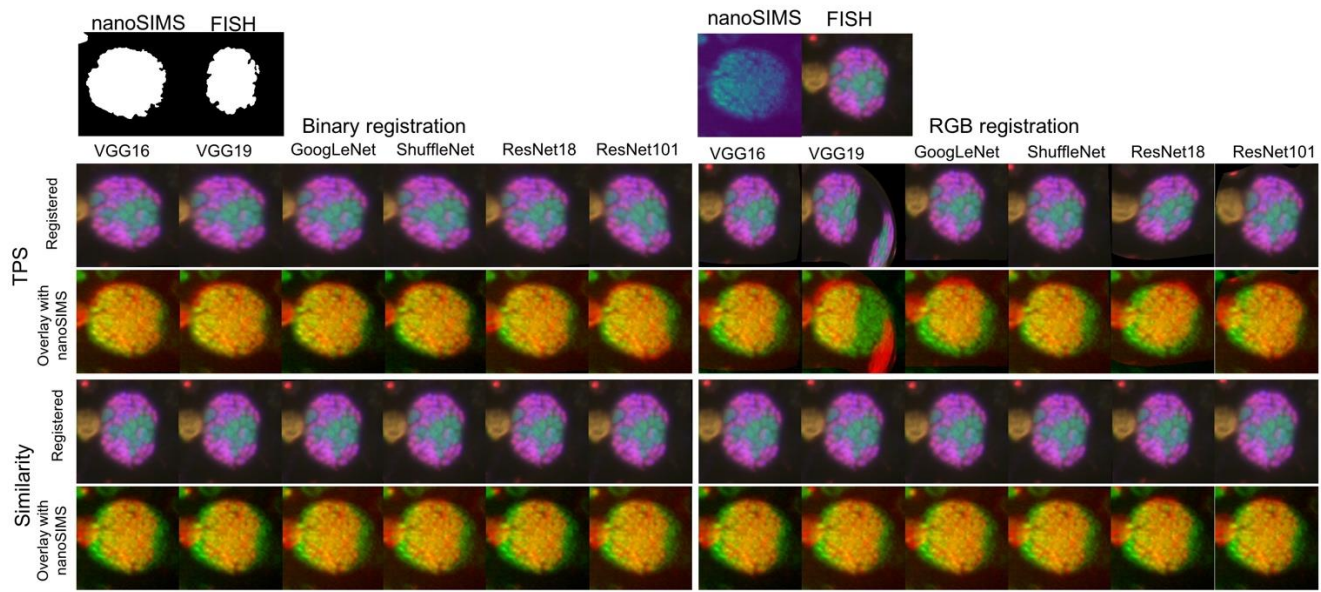

**Fig. 2.** Registration of a significantly deformed FISH image and nanoSIMS image using binary (left) and RGB (right) images as input with TPS-based (upper two panels) and similarity-based (lower two panels) schemes. Use grayscale image in print.

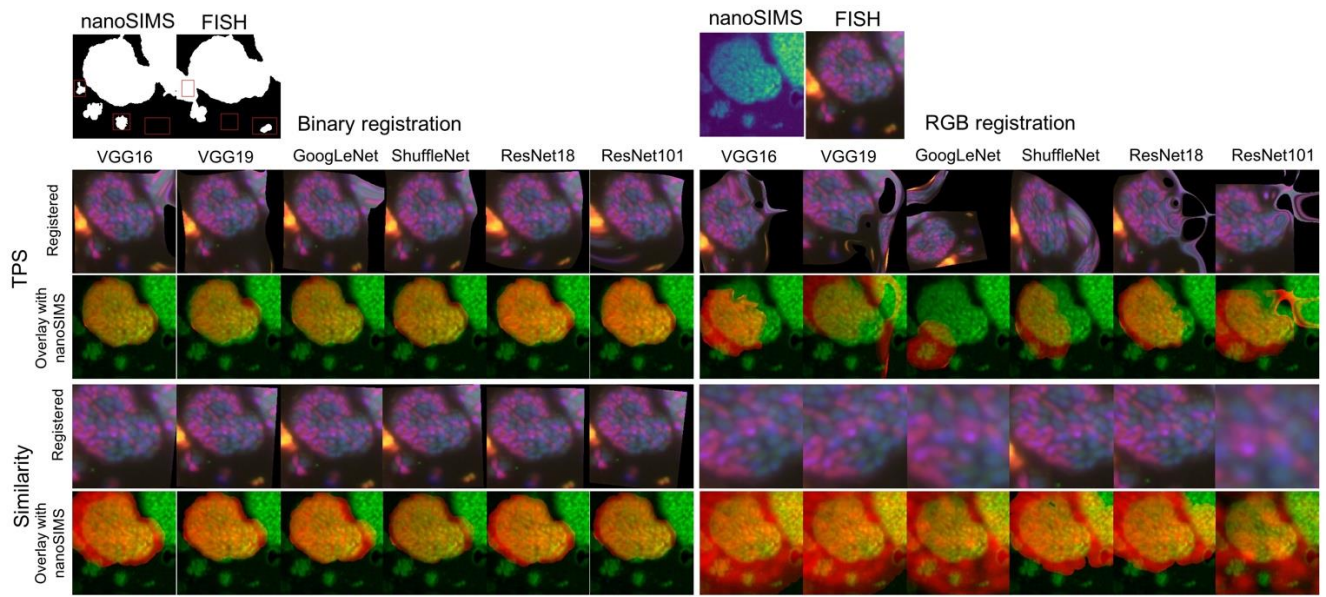

**Fig. 3.** Registration of a deformed FISH image with multiple components and nanoSIMS image using binary (left) and RGB (right) images as input with TPS-based (upper two panels) and similarity-based (lower two panels) schemes. Use grayscale image in print.

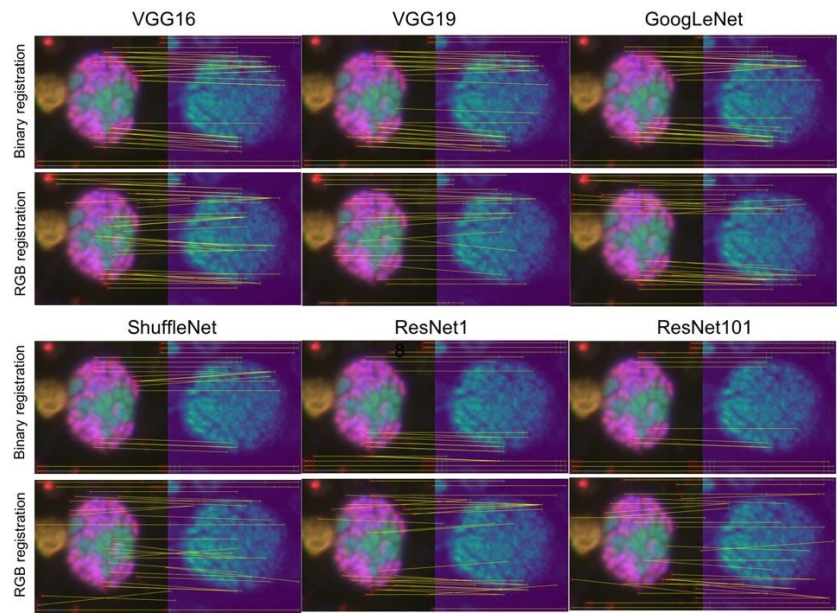

**Fig. 4.** Feature mapping after final thresholding during registration of a significantly deformed FISH image and nanoSIMS image using binary (upper two panels) and RGB inputs (lower two panels). Use grayscale image in print.

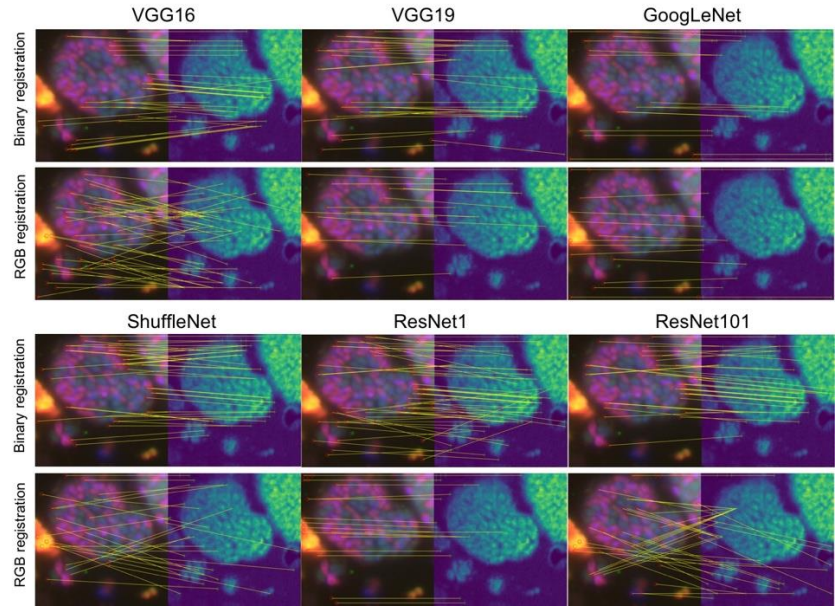

**Fig. 5.** Feature mapping after final thresholding during registration of a deformed FISH image with multiple components and nanoSIMS image using binary (upper two panels) and RGB inputs (lower two panels). Use grayscale image in print.

Table 1. Image registration accuracy for a significantly deformed FISH image and a deformed FISH image with multiple components. RMSE and AAID is smaller when the registration result is better, whereas SSIM is higher for a better aligned image. Best performance measures when compared across CNNs are highlighted in gray. Best performance for registration and CNN for a given image type are in bold. Best performance for each aggregate is indicated in red.

| | Significantly deformed | | | | | | | | | | | | Deformed with multiple components | | | | | | | | | | | |
|---|---|---|---|---|---|---|---|---|---|---|---|---|---|---|---|---|---|---|---|---|---|---|---|---|
| | Binary | | | | | | RGB | | | | | | Binary | | | | | | RGB | | | | | |
| | Similarity | | | TPS | | | Similarity | | | TPS | | | Similarity | | | TPS | | | Similarity | | | TPS | | |
| | RMSE | AAID | SSIM | RMSE | AAID | SSIM | RMSE | AAID | SSIM | RMSE | AAID | SSIM | RMSE | AAID | SSIM | RMSE | AAID | SSIM | RMSE | AAID | SSIM | RMSE | AAID | SSIM |
| GoogLeNet | 19.26 | 6.41 | 0.83 | 19.48 | 2.68 | 0.82 | 75.05 | 60.91 | 0.20 | 61.03 | 11.40 | 0.48 | 38.69 | **2.50** | 0.75 | **27.18** | 4.02 | 0.80 | 39.96 | 3.06 | 0.74 | 52.73 | 7.53 | 0.62 |
| ResNet101 | 19.66 | 4.93 | 0.814 | 13.97 | **1.66** | 0.88 | 75.07 | 63.37 | 0.21 | 49.53 | 19.77 | 0.53 | 38.78 | 2.56 | 0.74 | 31.15 | 4.56 | 0.78 | 38.81 | 4.28 | 0.74 | 45.07 | 6.57 | 0.71 |
| ResNet18 | 25.01 | 8.66 | 0.75 | **12.44** | 1.67 | **0.89** | 54.80 | 32.06 | 0.51 | **32.45** | **6.65** | **0.70** | 43.34 | 3.09 | 0.71 | 29.92 | 3.56 | 0.79 | 40.98 | 4.80 | 0.719 | 52.69 | 7.51 | 0.63 |
| ShuffleNet | 15.52 | 4.68 | 0.85 | 13.23 | 1.83 | 0.88 | 54.80 | 33.93 | 0.48 | 44.70 | 9.36 | 0.62 | 40.97 | 3.20 | 0.72 | 29.50 | 5.43 | 0.78 | 39.12 | **2.49** | 0.75 | **34.48** | 2.80 | **0.76** |
| VGG16 | 39.04 | 17.75 | 0.67 | 14.38 | 2.25 | 0.87 | 66.49 | 51.63 | 0.24 | 45.74 | 10.59 | 0.60 | 41.47 | 3.41 | 0.72 | 29.29 | 5.27 | 0.78 | 39.00 | 3.74 | 0.75 | 44.37 | 4.19 | 0.69 |
| VGG19 | 15.68 | 4.40 | 0.86 | 19.84 | 2.10 | 0.81 | 67.67 | 51.45 | 0.23 | 54.67 | 23.39 | 0.53 | 41.08 | 3.17 | 0.73 | 25.62 | 5.22 | **0.81** | 39.18 | 3.10 | 0.75 | 55.21 | 6.07 | 0.63 |

For validation purposes, we first compared our CNN-based methods to other well-recognized standard feature extraction methods SURF, KAZE, BRISK, Harris and FAST. None of them produced satisfactory results in our tests, which include registration of a modestly deformed (upper panel in Fig. S15), a significantly deformed (middle panel in Fig. S15), and a multiple-component deformed (lower panel in Fig. S15) FISH image with a nanoSIMS image. With RGB images as input, all standard feature extraction methods failed completely to register the FISH images with the nanoSIMS image due to the inherent shortcomings of the extracted and matched features.

To assess the quality of our CNN-based implementation, we compared it with the results of the state-of-the-art CoMIR method (40). Here we registered three distinct types of deformed FISH images: a moderately deformed FISH image (Fig. 6A), significantly deformed FISH image (Fig. 6B), and multiple-component deformed FISH image (Fig. 6C). CoMIR registered the FISH images with nanoSIMS images with high accuracy. Our proposed CNN-based methods performed comparably to the state-of-the-art CoMIR method, while significantly outperforming the rigid registration methods.

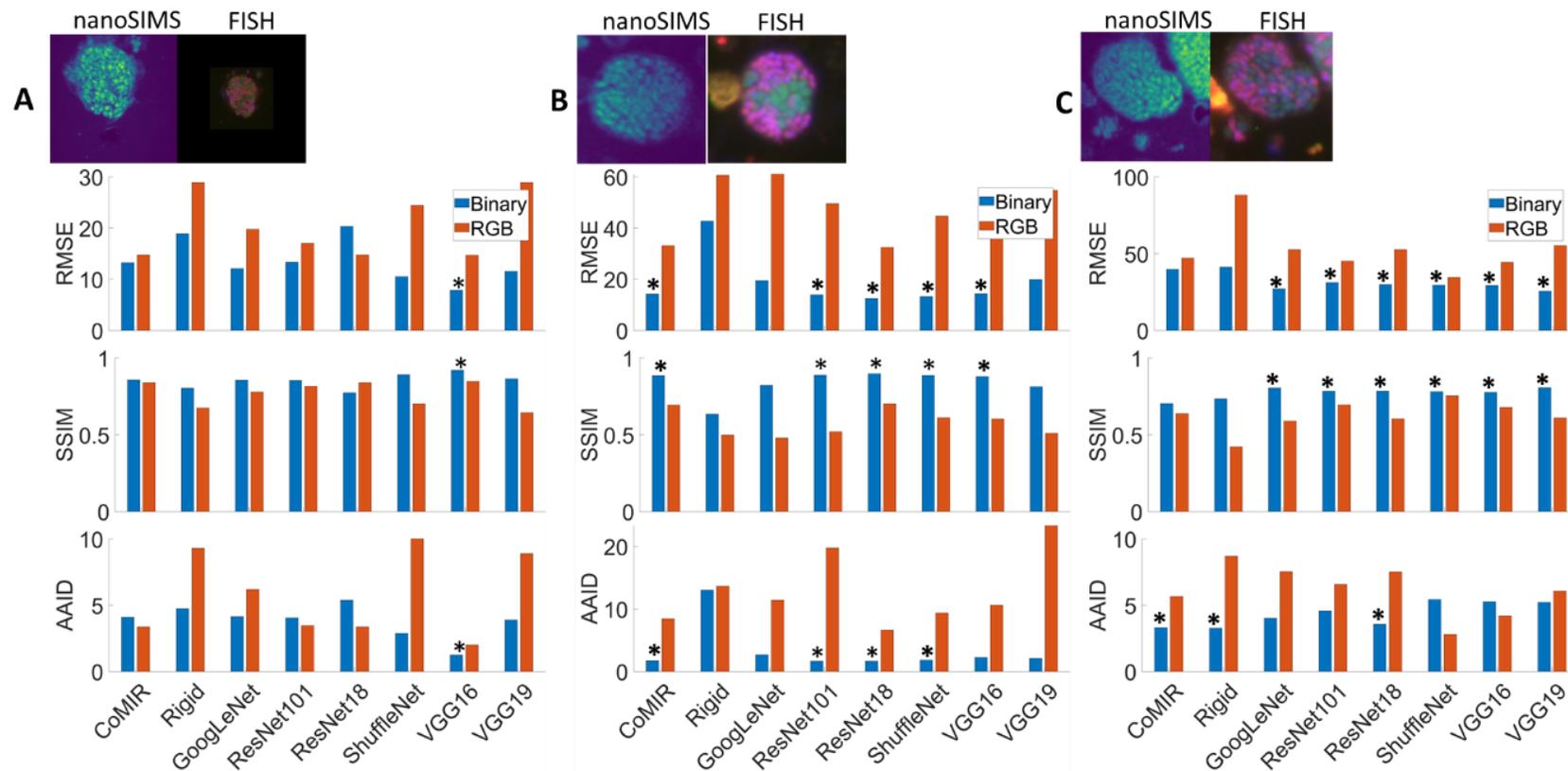

**Fig. 6.** Comparison of state-of-art method to our proposed CNN models. Image registration accuracy for a moderately deformed FISH image (A), significantly deformed FISH image (B), and deformed with multiple components (C): RMSE (upper panel), AAID (middle panel), and SSIM (lower panel). RMSE and AAID are smaller when the registration result is better, whereas SSIM is higher for a better aligned image. Use grayscale image in print.

## Conclusion

Our proposed workflow employed advanced CNN models to successfully extract shared feature points in FISH and nanoSIMS images for multimodal image registration. CNN-derived, feature-based non-rigid TPS registration methods significantly outperformed conventional similarity-based rigid registration methods, and produced registration results that were very comparable to those of the state-of-art method CoMIR method. We tested six CNN models using TPS-based non-rigid registration for different FISH and nanoSIMS images. The differences between the registration results obtained from the different CNN models considered in this study were minor. We demonstrated that image preprocessing with binarization is critical for final image registration and aggregate shape is a robust feature for microbiology-derived images such as FISH and nanoSIMS images. This may be largely owing to the significant differences in intra-aggregate patterns between the FISH and nanoSIMS images, leading to poor registration performance when using raw RGB images as input. It is also noted that images with significant background noise (non-ROI components) that cannot be easily removed via simple thresholding and binarization still pose a significant challenge. This highlights the importance of preserving aggregate morphology and reducing background noise in images with multiple aggregates.

## Acknowledgment

We thank Victoria Orphan and Gray Chadwick for providing the FISH and nanoSIMS images and the data on manual image registration used in (24). This work was supported by the U.S. Department of Energy, Office of Science, Office of Biological and Environmental Research, Genomic Science Program under Award Number DE-SC0016469 and DE-SC0020373 to CM.

Matlab scripts and images are made available at https://bitbucket.org/MeileLab/he_imageregistration/src/master/

**Disclosure statement**

The authors declare no conflict of interest.

**References**

1. Boxer SG, Kraft ML, Weber PK. Advances in imaging secondary ion mass spectrometry for biological samples. Annual Review of Biophysics. 2009;38:53-74.

2. Dekas AE, Connon SA, Chadwick GL, Trembath-Reichert E, Orphan VJ. Activity and interactions of methane seep microorganisms assessed by parallel transcription and FISH-NanoSIMS analyses. The ISME Journal. 2015;10:678–92.

3. Brown LG. A survey of image registration techniques. ACM Computing Surveys. 1992;24(4):325-76.

4. Heckbert PS. Fundamentals of texture mapping and image warping. Berkeley, CA: University of California, Berkeley; 1989.

5. Arad N, Reisfeld D. Image warping using few anchor points and radial functions. Computer Graphics Forum. 14: Blackwell Science Ltd; 1995. p. 35-46.

6. Hermessi H, Mourali O, Zagrouba E. Convolutional neural network-based multimodal image fusion via similarity learning in the shearlet domain. Neural Computing and Applications. 2018;30(7):2029-45.

7. Haskins G, Kruger U, Yan P. Deep learning in medical image registration: a survey. Machine Vision and Applications. 2020;31(1):8.

8. Zampieri A, Charpiat G, Girard N, Tarabalka Y, editors. Multimodal image alignment through a multiscale chain of neural networks with application to remote sensing. Proceedings of the European Conference on Computer Vision (ECCV); 2018.

9. Zhang H, Ni W, Yan W, Xiang D, Wu J, Yang X, et al. Registration of multimodal remote sensing image based on deep fully convolutional neural network. IEEE Journal of Selected Topics in Applied Earth Observations and Remote Sensing. 2019;12(8):3028-42.

10. Zhang M, Li W, Du Q. Diverse region-based CNN for hyperspectral image classification. IEEE Transactions on Image Processing. 2018;27(6):2623-34.

11. Han D, Liu Q, Fan W. A new image classification method using CNN transfer learning and web data augmentation. Expert Systems with Applications. 2018;95:43-56.

12. Kayalibay B, Jensen G, van der Smagt P. CNN-based segmentation of medical imaging data. arXiv preprint arXiv:170103056. 2017.

13. Bao S, Chung AC. Multi-scale structured CNN with label consistency for brain MR image segmentation. Computer Methods in Biomechanics and Biomedical Engineering: Imaging & Visualization. 2018;6(1):113-7.

14. Sokooti H, De Vos B, Berendsen F, Lelieveldt BP, Išgum I, Staring M, editors. Nonrigid image registration using multi-scale 3D convolutional neural networks. International Conference on Medical Image Computing and Computer-Assisted Intervention 2017: Springer.

15. Jiang P, Shackleford JA, editors. CNN driven sparse multi-level b-spline image registration. Proceedings of the IEEE Conference on Computer Vision and Pattern Recognition; 2018.

16. Uzunova H, Wilms M, Handels H, Ehrhardt J, editors. Training CNNs for image registration from few samples with model-based data augmentation. International Conference on Medical Image Computing and Computer-Assisted Intervention; 2017: Springer.

17. Ferrante E, Oktay O, Glocker B, Milone DH, editors. On the adaptability of unsupervised CNN-based deformable image registration to unseen image domains. International Workshop on Machine Learning in Medical Imaging; 2018: Springer.

18. Lee JA, Liu P, Cheng J, Fu H. A deep step pattern representation for multimodal retinal image registration. Proceedings of the IEEE/CVF International Conference on Computer Vision. 2019:5077-86.

19. Hu J, Luo Z, Wang X, Sun S, Yin Y, Cao K, et al. End-to-end multimodal image registration via reinforcement learning. Medical Image Analysis. 2021(68):101878.

20. Hering A, Häger S, Moltz J, Lessmann N, Heldmann S, van Ginneken B. CNN-based Lung CT Registration with Multiple Anatomical Constraints. Medical Image Analysis. 2021:102139.

21. Boveiri HR, Khayami R, Javidan R, Mehdizadeh A. Medical image registration using deep neural networks: A comprehensive review. Computers & Electrical Engineering. 2020;87:106767.

22. Szegedy C, Liu W, Jia Y, Sermanet P, Reed S, Anguelov D, et al., editors. Going deeper with convolutions. Proceedings of the IEEE conference on computer vision and pattern recognition; 2015.

23. Simonyan K, Zisserman A. Very deep convolutional networks for large-scale image recognition. 3rd International Conference on Learning Representations; San Diego, CA, USA2015.

24. McGlynn SE, Chadwick GL, Kempes CP, Orphan VJ. Single cell activity reveals direct electron transfer in methanotrophic consortia. Nature. 2015;526(7574):531-5.

25. Zhang X, Zhou X, Lin M, Sun J, editors. ShuffleNet: An Extremely Efficient Convolutional Neural Network for Mobile Devices. Proceedings of the IEEE conference on computer vision and pattern recognition; 2018.

26. He K, Zhang X, Ren S, Sun J, editors. Deep residual learning for image recognition. Proceedings of the IEEE conference on computer vision and pattern recognition; 2016.

27. Belongie S, Malik J, editors. Matching with Shape Contexts. 2000 Proceedings Workshop on Content-based Access of Image and Video Libraries; 2000; Head Island, SC, USA.

28. Jonker R, Volgenant A. A shortest augmenting path algorithm for dense and sparse linear assignment problems. Computing. 1987;38(4):325-40.

29. Powell MJD. A thin plate spline method for mapping curves into curves in two dimensions. Computational Techniques and Applications (CTAC '95). 1995.

30. Goshtasby A. Image registration by local approximation methods. Image and Vision Computing. 1988;6:255-61.

31. Goshtasby A. Piecewise linear mapping functions for image registration. Pattern Recognition. 1986;19:459-66.

32. Zhang Z, Han D, Dezert J, Yang Y. A new image registration algorithm based on evidential reasoning. Sensors. 2019;19(5):1091.

33. Goshtasby A. 2-D and 3-D Image Registration for Medical, Remote Sensing, and Industrial Applications: John Wiley & Sons; 2005.

34. Sprengel R, Rohr K, Stiehl HS. Thin-plate spline approximation for image registration. Proceedings of 18th Annual International Conference of the IEEE Engineering in Medicine and Biology Society. 1996;3:1190-1.

35. Bay H, Ess A, Tuytelaars T, Van Gool L. SURF:Speeded Up Robust Features. Computer Vision and Image Understanding (CVIU). 2008;110(3):346–59.

36. Alcantarilla PF, Bartoli A, Davison AJ. KAZE Features. In: Fitzgibbon A, Lazebnik S, Perona P, Sato Y, Schmid C, editors. Computer Vision – ECCV 2012 Lecture Notes in Computer Science. Berlin, Heidelberg: Springer; 2012. p. 7577.

37. Leutenegger S, Chli M, Siegwart R, editors. BRISK: Binary Robust Invariant Scalable Keypoints. 2011 International Conference on Computer Vision; 2011; Barcelona, Spain.

38. Harris C, Stephens M. A Combined Corner and Edge Detector. Proceedings of the 4th Alvey Vision Conference. 1988:147-51.

39. Rosten E, Drummond T. Fusing Points and Lines for High Performance Tracking. Proceedings of the IEEE International Conference on Computer Vision. 2005;2:1508–11.

40. Wetzer E, Pielawski N, Öfverstedt J, Lu J, Wählby C, Lindblad J, et al. CoMIR: Contrastive multimodal image representation for registration. 34th Conference on Neural Information Processing Systems (NeurIPS 2020); Vancouver, Canada.2021.